\documentclass[a4paper,twoside]{article}

\usepackage[pdftex]{graphicx}
\usepackage[tight,footnotesize]{subfigure}
\usepackage{caption}
\usepackage{url}
\usepackage{subcaption}
\usepackage{calc}
\usepackage{cite}
\usepackage{amssymb}
\usepackage{amstext}
\usepackage{amsmath}
\usepackage{amsthm}
\usepackage{multicol}
\usepackage{pslatex}
\usepackage{apalike}
\usepackage{algorithm2e}
\usepackage[bottom]{footmisc}
\usepackage{SCITEPRESS}     

\begin{document}

\title{PatchSVD: A Non-uniform SVD-based Image Compression Algorithm}

\author{\authorname{Zahra Golpayegani\sup{1}, Nizar Bouguila\sup{2}
}
\affiliation{Gina Cody School of Engineering, Concordia University, Montreal, Canada
\email{\sup{1}zahra.golpayegani@mail.concordia.ca,\sup{2}nizar.bouguila@concordia.ca}}
}


\abstract{Storing data is particularly a challenge when dealing with image data which often involves large file sizes due to the high resolution and complexity of images. Efficient image compression algorithms are crucial to better manage data storage costs. In this paper, we propose a novel region-based lossy image compression technique, called PatchSVD, based on the Singular Value Decomposition (SVD) algorithm. We show through experiments that PatchSVD outperforms SVD-based image compression with respect to three popular image compression metrics. Moreover, we compare PatchSVD compression artifacts with those of Joint Photographic Experts Group (JPEG) and SVD-based image compression and illustrate some cases where PatchSVD compression artifacts are preferable compared to JPEG and SVD artifacts.}

\onecolumn \maketitle \normalsize \setcounter{footnote}{0} \vfill



\section{\uppercase{Introduction}}
\label{sec:introduction}

Enormous amounts of data are generated every day by various sources, including social media, sensors on wearable devices, and smart gadgets. In many cases, the data needs to be stored on a small device to serve a specific purpose. For instance, a real-time defect detector stores images of its camera feed and sends them to a lightweight model to catch possible flaws in a production line \cite{pham2023yolo}. Such small devices are limited in storage capacities; therefore, it is essential to design efficient data storage algorithms capable of storing the data without significant loss in quality.

\begin{figure}
    \centering
    \includegraphics[width=\columnwidth]{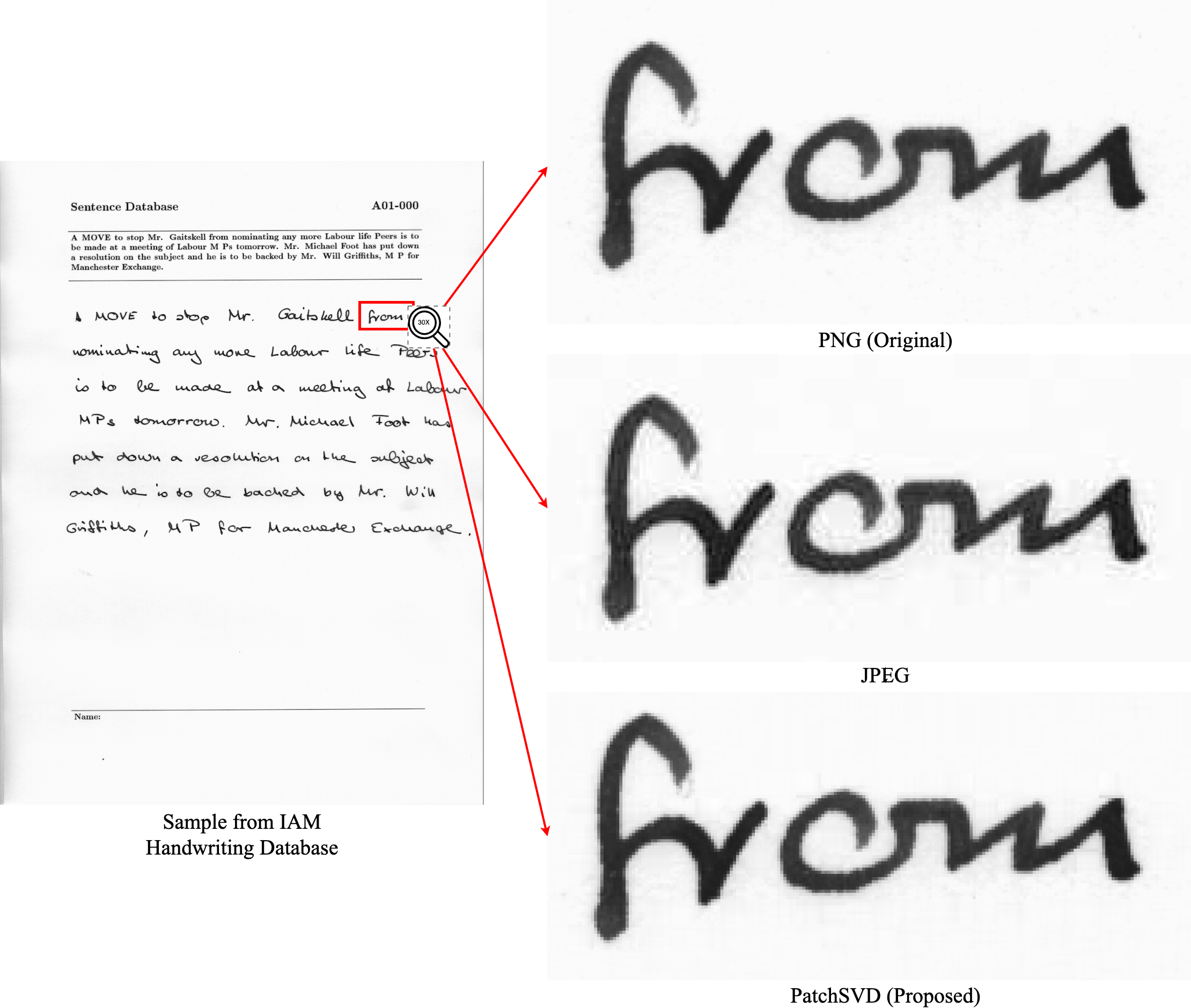}
    \caption{JPEG produces more compression artifacts in images containing text compared to the proposed method. The sample is taken from the IAM Handwriting Database and the image has been zoomed in 30 times the original size to better visualize the compression artifacts.}
    \label{fig:from}
\end{figure}

Images are one of the most commonly used data formats, and storing an image file on a digital device can require anywhere from kilobytes to megabytes of storage space, depending on the complexity of the image and the storage technique. Image compression algorithms can be categorized into lossless and lossy compression methods. In lossless methods, such as Portable Network Graphics (PNG) image compression, the original input can be reconstructed after compressing because compression was achieved by removing statistical redundancies. However, only low compression ratios are achievable through lossless image compression, and image compression is not always guaranteed. On the other hand, lossy compression techniques, such as Joint Photographic Experts Group (JPEG) compression \cite{wallace1991jpeg}, achieve higher compression ratios by allowing more information loss, but the reconstructed image sometimes contains visible compression artifacts. Specifically, JPEG image compression is based on the assumption that within an $8 \times 8$ pixel block, there are no sharp changes in intensity. However, in some use cases, such as compressing images of text or electronic circuit diagrams, this assumption does not hold, and JPEG creates visible compression artifacts around the drawn lines. Figure \ref{fig:from} compares the JPEG compression artifacts with those of the proposed method presented in this paper, using an example image containing text sourced from the IAM Handwriting Database \cite{marti2002iam}.

\begin{figure*}
  \centering

  \subfigure[Original image]{
    \includegraphics[width=0.22\linewidth]{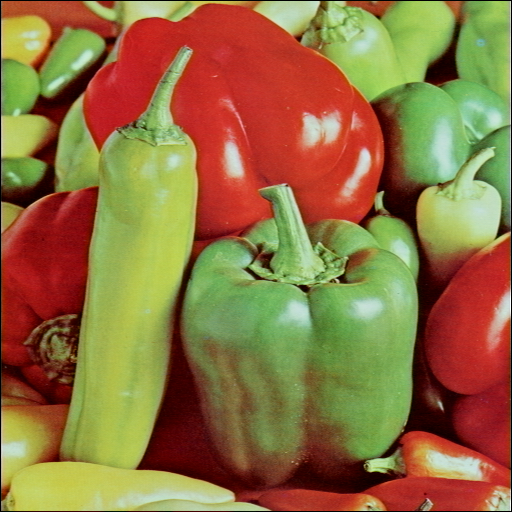}
    \label{subfig:a}
  }
  \subfigure[Difference image ($\Delta$)]{
    \includegraphics[width=0.22\linewidth]{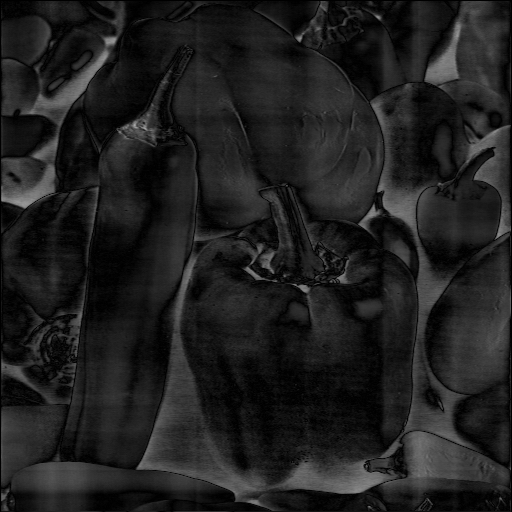}
    \label{subfig:b}
  }
  \subfigure[Complex (in gray) and simple (in black) patches]{
    \includegraphics[width=0.22\linewidth]{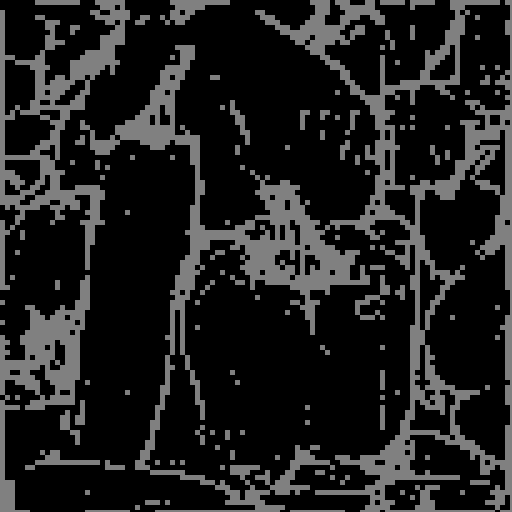}
    \label{subfig:c}
  }
  \subfigure[PatchSVD output]{
    \includegraphics[width=0.22\linewidth]{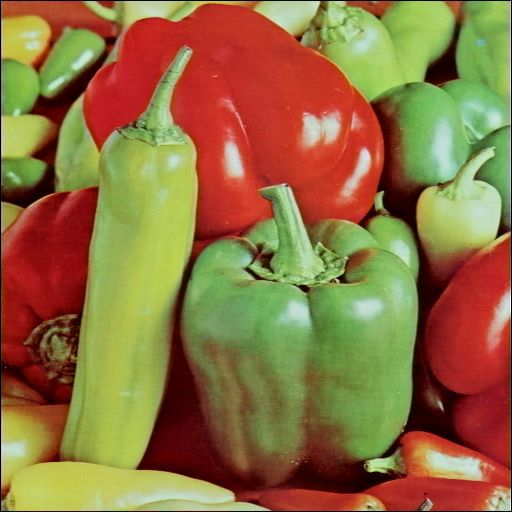}
    \label{subfig:d}
  }
  \caption{PatchSVD algorithm first applies low-rank SVD to the original image and subtracts the low-rank approximation from the original image (Figure \ref{subfig:a}) to obtain $\Delta$ (Figure \ref{subfig:b}). Then, by applying a score function, PatchSVD calculates the patches that contain more information according to $\Delta$ (\ref{subfig:c}) to create the final compressed image (see Figure \ref{subfig:d}).}
  \label{fig:overall}
\end{figure*}

To improve the compression results, some traditional methods compress regions of interest (ROIs) with a higher bit-rate than the rest of the regions \cite{christopoulos2000jpeg2000}, however, the ROIs are not detected automatically and should be specified by the user. Deep learning-based approaches have also been applied to image compression \cite{cheng2018deep,toderici2017full,agustsson2017soft,li2018learning,cheng2019deep,prakash2017semantic}. Nevertheless, training deep learning models require high computational power and during inference, the model has to exist on the device that runs the compression code, which calls for additional storage compared to traditional methods.

In this paper, we propose a new image compression algorithm based on Singular Value Decomposition (SVD), called PatchSVD\footnote{PatchSVD source code is available at \url{https://github.com/zahragolpa/PatchSVD}}. First, we explain how PatchSVD works, what is the compression ratio we can achieve using PatchSVD, and what are the required conditions to compress an image using PatchSVD. Then, through extensive experiments, we demonstrate that PatchSVD outperforms SVD in terms of Structural Similarity Index Measure (SSIM), Peak Signal-to-Noise Ratio (PSNR), and Mean Squared Error (MSE) metrics. Moreover, through examples, we show that PatchSVD is more robust against sharp pixel intensity changes, therefore, it is preferable over JPEG in some use cases, including compressing images of text, where high changes in pixel intensity exist.

\section{\uppercase{Preliminaries}}
In this section, we briefly overview SVD and its basic properties.
Singular Value Decomposition or SVD is an algorithm that factorizes a given matrix 
$A \in \mathbb{R}^{m \times n}$
into three components $A = USV^T$, where $U_{m \times m}$ and $V^T_{n \times n}$ are orthogonal matrices and $S_{m \times n}$ is a diagonal matrix containing singular values in descending order. The singular values are non-negative real numbers that represent the magnitude of the singular vectors; i.e., larger singular values indicate directions in the data space where there is more variability. Therefore, by keeping the larger singular values and their corresponding vectors, the original matrix can be approximated with minimum information loss.

The maximum number of linearly independent rows or columns of a matrix is called the rank ($r$) of that matrix. When a matrix $A$ is decomposed using SVD, the number of non-zero elements on the diagonal of $S$ is equal to $r$. By retaining only the first $r$ elements from $U$, $S$, and $V^T$ we get an $r$-rank approximation $A_r = U_rS_rV_r^T$ for the matrix $A$.

If we use $k$-rank SVD to compress an image represented by matrix $A_{m \times n}$, the number of elements we need to store to represent the compressed image is calculated by summing up the number of elements from each SVD component. Therefore, storing the $k$-rank version of $A_{m \times n}$ with $m \times n$ elements only requires $S_{SVD} = k(m + n + 1)$ values. Note that the $k$-rank approximation of matrix $A$ keeps most of the useful information about the image while reducing the storage requirements for the input, especially if the rows and columns in the image are highly correlated ($r \ll min(m, n)$). However, when higher compression ratios are required, it is beneficial to sacrifice more information to save more storage, which is achievable by choosing $k_{SVD} < k$.


\section{\uppercase{Related Works}}
SVD has been used before in literature for image compression \cite{andrews1976singular,akritas2004applications,kahu2013image,prasantha2007image,tian2005investigation,cao2006singular}. In \cite{ranade2007variation}, a variation on image compression using SVD has been proposed that applies a data-independent permutation on the input image before performing SVD-based image compression as a preprocessing step. In \cite{DBLP:journals/corr/Sadek12}, a forensic tool is developed using SVD that embeds a watermark into the image noise subspace. Another application of SVD applied to image data is image recovery discussed in \cite{chen2018singular} where authors used SVD for matrix completion.

Few studies have investigated region-based or patch-based image compression using techniques similar to SVD. In \cite{lim2014gui}, a GUI system is designed that takes Regions of Interest (ROI) in medical images to ensure near-zero information loss in those regions compared to the rest of the image when compressed. However, users need to select the ROIs manually, and Principle Component Analysis (PCA) is used instead of SVD. \cite{lim2022region} used a patch-based PCA algorithm that eliminates the need to manually select the ROIs in brain MRI scans using the brain symmetrical property. However, their approach has very limited use cases because of the assumed symmetrical property in the images.


Joint Photographic Experts Group (JPEG) compression \cite{wallace1991jpeg} is one of the most commonly used image compression methods, applicable to both grayscale and color continuous-tone images. JPEG works by transforming an image from the spatial domain to the frequency domain using Discrete Cosine Transforms (DCTs) \cite{ahmed1974discrete}. Each image is divided into $8 \times 8$ pixels blocks, transformed using DCT, and then quantized according to a quantization table, followed by further compression using an entropy encoding algorithm, such as Huffman coding \cite{huffman1952method}. While JPEG is used in many use cases, it fails to perform well in examples where sudden changes in intensity exist within the blocks. 

We aim to extend the previous works by automatically selecting the complex patches and utilizing SVD with respect to the image context to compress images with minimum information loss and achieve significant reductions in storage without any training required using simple mathematical operations. 

\begin{figure*}
  \centering
  \includegraphics[width=\linewidth]{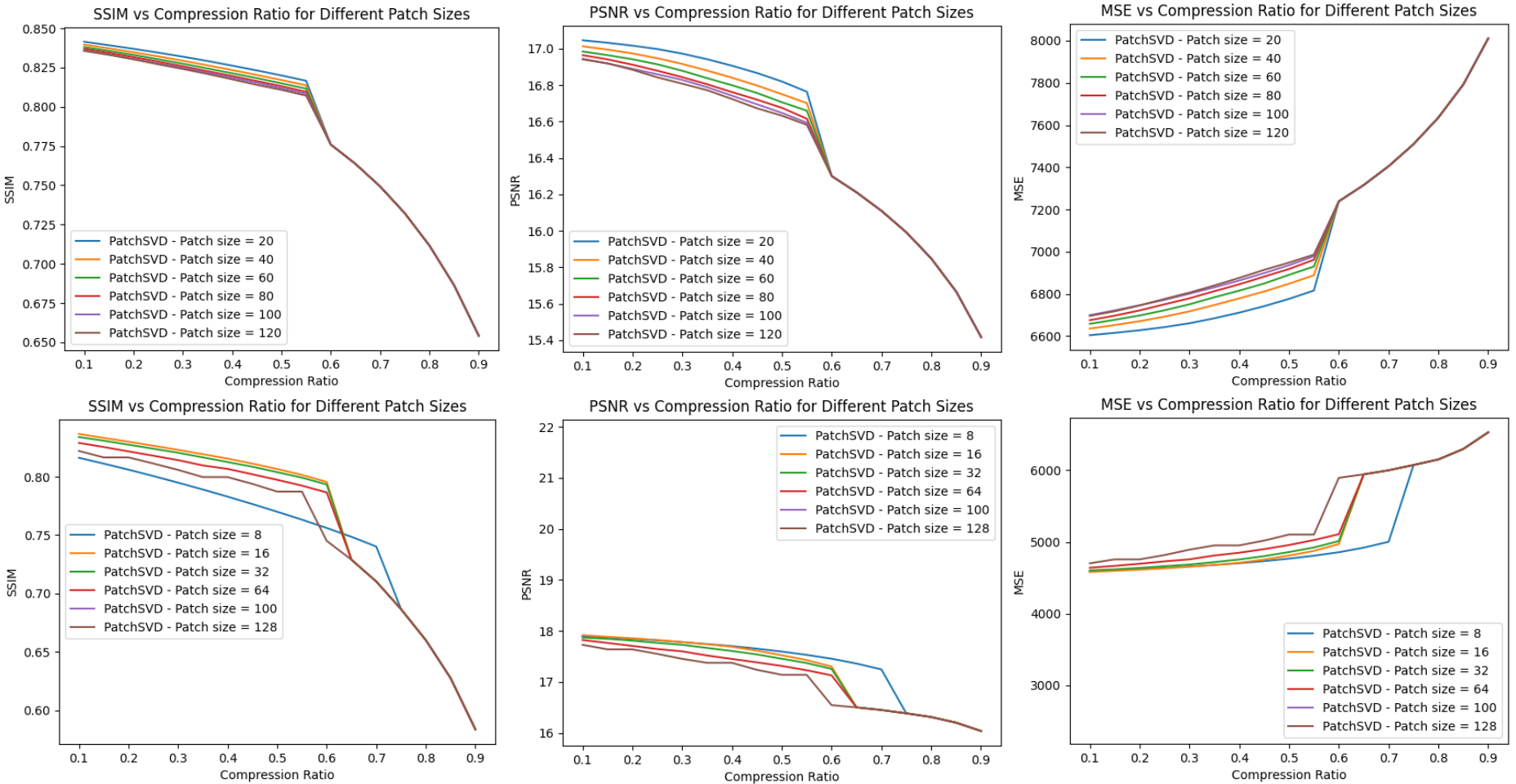}
  
  \caption{This figure shows the effect of patch size on the performance of the PatchSVD algorithm across CLIC (first row) and Kodak (second row) datasets.}
  \label{fig:various-patch-sizes}
\end{figure*}

\section{\uppercase{Method}}
\subsection{PatchSVD Algorithm}
To compress $A_{m \times n}$ using PatchSVD, we first compute the SVD of the image and take the first $k_{SVD}$ singular values to reconstruct the $k_{SVD}$-rank approximation. Then, we subtract the $k_{SVD}$-rank approximation from the original image to get matrix $\Delta$ (see Figure \ref{subfig:b}). In our experiments, we selected the initial $k_{SVD} = 1$. Based on the properties of SVD, the $k_{SVD}$-rank approximation captures most of the image information; therefore, high values in $\Delta$ (white pixels) indicate pixels that were not captured by the first singular values, and low values in $\Delta$ (black pixels) would give us those locations were almost all the information in the original image was captured by the $k_{SVD}$-rank approximation. Therefore, to distinguish the complex patches that SVD missed from the simpler ones, we investigate the values in $\Delta$ and determine if a patch is complex or simple (see Figure \ref{subfig:c}). In other words, the $\Delta$ matrix helps us find the areas that would introduce large compression errors if we used the standard SVD for image compression. We utilize $\Delta$ as a heuristic function to minimize the compression error by applying non-uniform compression.

More specifically, we split the image corresponding to the $\Delta$ matrix into patches of size $P_x \times P_y$. If the image is not divisible by the patch dimensions, we add temporary margins to the sides with pixel values equal to the average pixel value in the image. Then, we loop over the patches and assign a score to each patch according to a score function. Next, we sort the patches based on the score and select the top $n_c$ complex patches. The number of complex patches is determined based on the desired compression ratio, following Equation \ref{eq:calc_n_c}.

After we find the complex patches, we run SVD for each patch and take the $k$-rank approximation with two different constants; $k_c$ for complex patches and $k_s$ for simple patches where $k_s \leq k_c$. When we encounter the patches that contain the extra margin, we remove the margin before performing $k$-rank approximation using SVD. Finally, we put the compressed patches together to form the compressed image (see Figure \ref{subfig:d}). PatchSVD algorithm is described in detail in Algorithm \ref{alg:patchsvd}. We argue that PatchSVD is a more flexible image compression algorithm compared to SVD and JPEG because it allows you to assign non-uniform importance to each patch in the image according to a customizable function. While our method relies on the 1-rank SVD to calculate the $\Delta$ matrix and employs the standard deviation score function to sort patches, it is noteworthy that various techniques, such as graph-based approaches, gradient-based methods, edge detection, and expert knowledge, can also be employed to detect and sort complex patches. The choice may depend on the specific requirements of the application.



\begin{algorithm}
  \caption{PatchSVD Image Compression}
  \label{alg:patchsvd}

  \KwIn{$A_{m \times n}, k_s, k_c, P_x, P_y, \texttt{CR}$}
  \KwOut{\texttt{cmpr\_image}}

  $\frac{n_c}{t} \gets (\frac{P_xP_y (1 - \texttt{CR})}{P_x + P_y + 1} - k_s)\frac{1}{k_c - k_s}$\;
  $\Delta \gets A - \mathrm{k\_rank\_SVD}(A, 1)$\;
  $\Delta_{\texttt{patches}} \gets \mathrm{patch\_and\_add\_margin}(\Delta, P_x, P_y)$\;

  \If{$\frac{n_c}{t} \times \texttt{num\_patches} < 1$}{
    $\texttt{k\_SVD} \gets \texttt{int}((1 - \texttt{CR}) \times \frac{m \times n}{m + n + 1})$\;
    \textbf{return} $\mathrm{k\_rank\_SVD}(A, \texttt{k\_SVD})$\;
  }

  \For{$\texttt{patch} \text{ in } \Delta_{\texttt{patches}}$}{
    $\Delta_{\texttt{scores}}[\texttt{patch}] \gets \mathrm{score}(\texttt{patch})$\;
  }

  $\Delta_{\texttt{scores}} \gets \mathrm{sort}(\Delta_{\texttt{scores}})$\;

  \For{$\texttt{patch} \text{ in } \Delta_{\texttt{patches}}$}{
    $U, S, V^t \gets \mathrm{SVD}(\texttt{patch})$\;
    \If{$\mathrm{index}(\texttt{patch}) \leq n_c$}{
      $k \gets k_c$\;
    }
    \Else{
      $k \gets k_s$\;
    }
    $\texttt{cmpr\_patch} \gets \mathrm{k\_rank\_SVD}(\texttt{patch}, k)$\;
    $\texttt{cmpr\_patches}.\mathrm{append}(\texttt{cmpr\_patch})$\;
  }

  $\texttt{cmpr\_image} \gets \mathrm{arrange}(\texttt{cmpr\_patches})$\;

  \textbf{return} $\texttt{cmpr\_image}$\;

\end{algorithm}

\begin{figure*}

  \centering
  \begin{minipage}[b]{\textwidth}
    \centering
    \includegraphics[width=\textwidth]{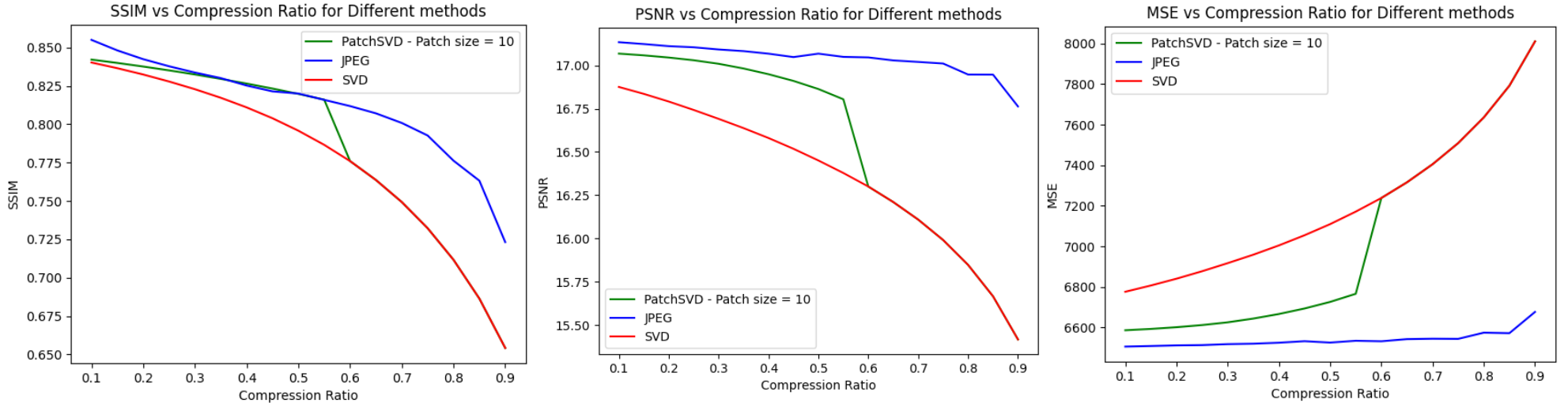}
  \end{minipage}
  \begin{minipage}[b]{\textwidth}
    \centering
    \includegraphics[width=\textwidth]{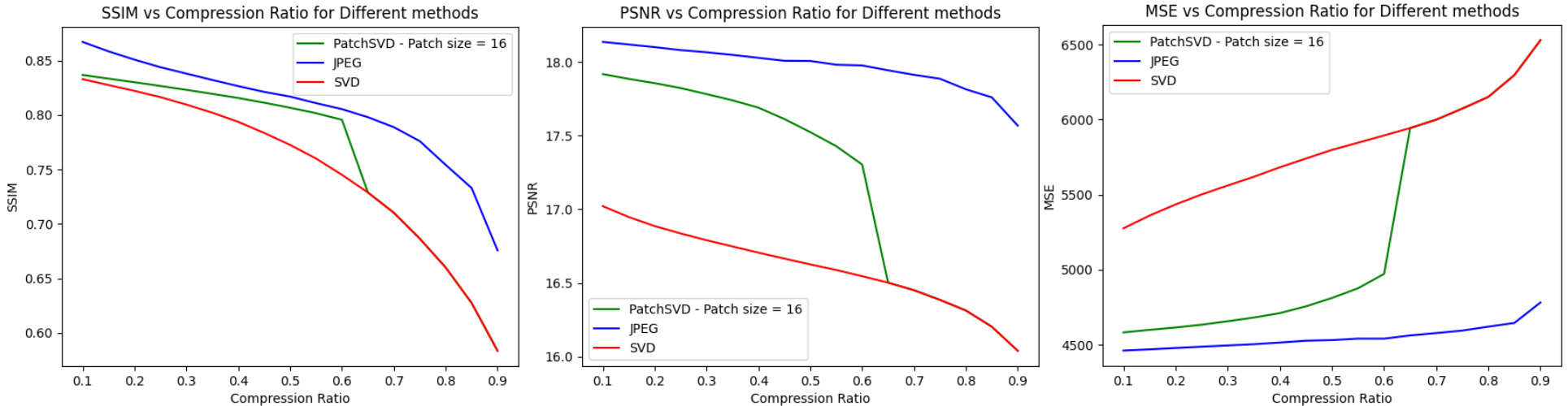}
  \end{minipage}

  
  \caption{This figure compares SSIM, PSNR, and MSE metrics for PatchSVD, JPEG, and SVD image compression algorithms on the CLIC dataset with a patch size of 10 (top row) and the Kodak dataset with a patch size of 16 (bottom row).}
  \label{fig:performance-kodak}
\end{figure*}

\subsection{Compression Ratio (CR)}
Suppose we want to compress an input image of size $m \times n$ using PatchSVD and we choose the patch size to be $P_x \times P_y$. The amount of compression we get depends on the values we choose for $k_c$ and $k_s$, i.e., lower-rank approximations result in higher amounts of image compression. More specifically, the number of digits we need to save, $S_{\text{PatchSVD}}$, to represent an image using PatchSVD is $S_{\text{PatchSVD}} = n_ck_c(P_x + P_y + 1) + n_sk_s(P_x + P_y + 1)$, where $n_c$ and $n_s$ are the number of complex and simple patches, respectively. Taking $k_c = k_s = k_{\text{SVD}}$ would result in the storage required by the SVD algorithm ($S_{\text{SVD}}$) which is equivalent to the storage needed by the PatchSVD algorithm when $P_x = n$, $P_y = m$, $n_c = 1$, and $n_s = 0$ which is equal to $S_{\text{SVD}} = k_{\text{SVD}}(m + n + 1)$.

Therefore, the Compression Ratio (CR) can be calculated using the following formula:

\begin{equation}
\label{eq:compression_ratio}
\begin{gathered}
     \text{CR} = \frac{S_{\text{Original}} - S_{\text{PatchSVD}}}{S_{\text{Original}}} \\
    = 1 - \frac{(P_x + P_y + 1)(n_c k_c + n_s k_s)}{P_xP_y(n_c + n_s)}
\end{gathered}
\end{equation}

where $S_{\text{Original}}$ is the storage needed for the original image. From \ref{eq:compression_ratio}, we can calculate the number of complex patches $n_c$ when the required compression ratio is known:

\begin{equation}
\label{eq:calc_n_c}
\begin{gathered}
    \text{CR} = 1 - \frac{(P_x + P_y + 1)(n_c k_c + n_s k_s)}{P_xP_y(n_c + n_s)} \\
    \Longrightarrow{} \frac{n_c}{t} = (\frac{P_xP_y (1 - CR)}{P_x + P_y + 1} - k_s)\frac{1}{k_c - k_s}
\end{gathered}
\end{equation}


  

where $t$ is the total number of patches in the image. We need to ensure $CR \geq 0$ to have a compressed image; therefore, the following should hold:

\begin{equation}
\label{eq:compression_condition}
    0 \leq \frac{n_c}{n_c + n_k} \leq \frac{\frac{P_xP_y}{P_x + P_y + 1} - k_s}{k_c - k_s}   
\end{equation}

which requires two conditions to be true:
\begin{equation}
\label{eq:condition}
\frac{P_xP_y}{P_x + P_y + 1} \geq k_s    
\end{equation}

and

\begin{equation}
k_c \geq k_s
\end{equation}

Moreover, we can observe from the condition in \ref{eq:compression_condition} that there is a trade-off between the proportion of the complex patches and the values we choose for $k_c$ and $k_s$. For instance, if we want to keep a higher proportion of complex patches, the difference between $k_c$ and $k_s$ should be higher and $k_s$ should be much smaller than $k_c$. Note that since $k_c \geq k_s \geq 0$, the proportion of complex patches cannot be higher than a threshold. On the other hand, if the proportion is too small, the number of complex patches may end up being less than 1, in which case PatchSVD will fall back to the standard SVD-based image compression algorithm.



\begin{figure*}
  \centering
  \subfigure[Image compression at CR = 85\% applied to "kodim06.png" from the Kodak dataset.]{
    \includegraphics[width=0.9\linewidth]{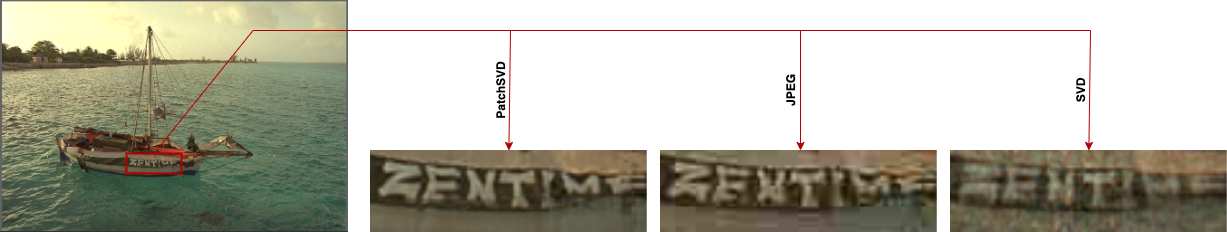}
    \label{subfig:artifact-a}
  }
  \subfigure[Image compression at CR = 20\% applied to "kodim09.png" from the Kodak dataset.]{
    \includegraphics[width=0.9\linewidth]{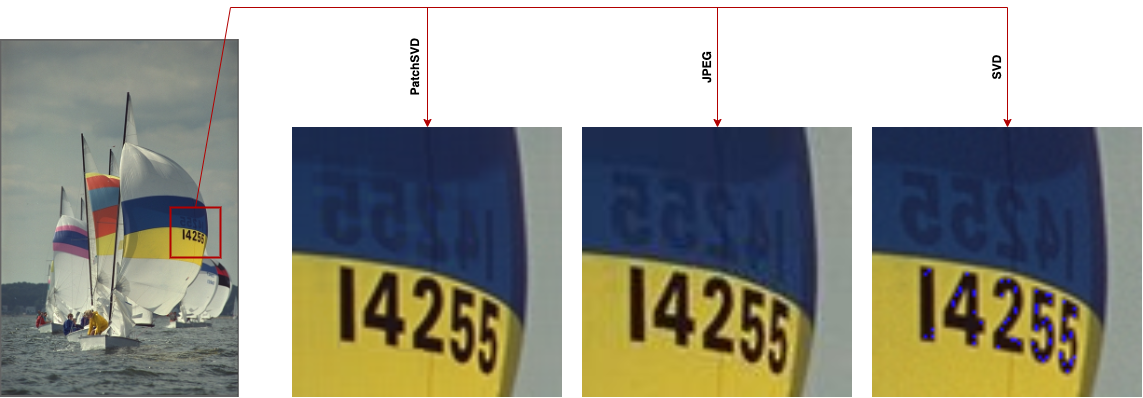}
    \label{subfig:artifact-b}
  }
  
  \captionsetup{width=0.9\linewidth}
  \caption{PatchSVD, JPEG, and SVD compression algorithms applied to two image samples from the Kodim dataset show the compression artifacts produced by each algorithm. As you can see, PatchSVD produces more sharp edges which results in perfectly legible text even after the image has been greatly compressed. }
  \label{fig:artifacts}
\end{figure*}

\subsection{Patch Size Lower Bound}
Not every arbitrary patch size works with the PatchSVD algorithm. While the image size is a trivial upper bound for $P$, it should be noted that greater patch sizes will result in less accurate scoring. More specifically, if the patch size is too large, the score function will lose its sensitivity to complex versus simple patches because of its less local domain. To find a lower bound for patch size, we simplify Equation \ref{eq:condition} by assuming that we are using a square patch with $P_x = P_y = P$. Then, we will have $\frac{P^2}{2P + 1} \geq k_s$.

By simplifying this inequality further, we get a lower bound on the patch size which is $P \geq k_s + \sqrt{k_s^2 + k_s} \geq 0$.

\section{\uppercase{Experiments}}

With PatchSVD, we compress images from two datasets using different patch sizes to evaluate the effect of patch size on the performance of the algorithm. Then, we pick the best patch size for each dataset and compare PatchSVD with SVD and JPEG according to three metrics and by visually comparing the compression artifacts. We also briefly discuss some choices for PatchSVD score functions and how they compare with each other.

\subsection{Datasets and Metrics}
We evaluated our method on Kodak \cite{kodak-suite} and CLIC \cite{CLIC2020} datasets because they contain original PNG format images that were never converted to JPEG. Kodak is a classic dataset that is frequently used for evaluating image compression algorithms. It contains 24 full-color (24 bits per pixel) images that are either 768x512 or 512x768 pixels large. The images in this dataset capture a variety of lighting conditions and contain different subjects. The CLIC dataset was introduced in the lossy image compression track for the Challenge on Learned Image Compression in 2020 and includes both RGB and grayscale images. For all the experiments, we utilized the test split, which comprises 428 samples.

We use traditional image compression metrics, including Mean Squared Error (MSE), Peak Signal-to-noise Ratio (PSNR), and Structural Similarity Index Measure (SSIM) \cite{wang2004image} to evaluate the image compression performance of each method. Lower MSE means better performance, indicating reduced pixel deviations. Higher PSNR and SSIM values signal superior image quality with less distortion and increased similarity between original and processed images.






\begin{figure*}
  \centering
  \begin{minipage}[b]{\textwidth}
    \centering
    \includegraphics[width=\textwidth]{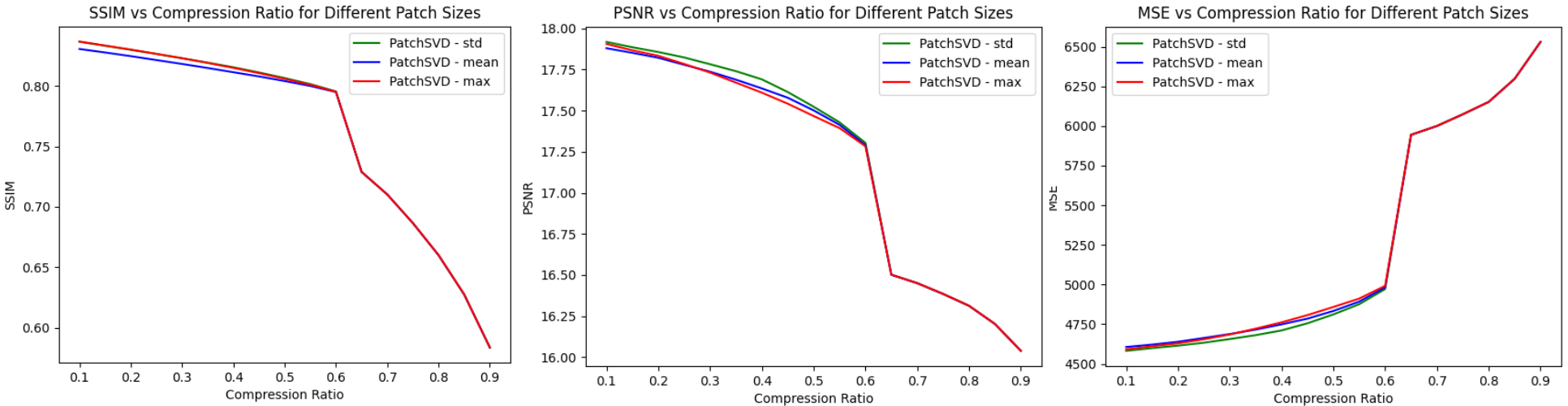}
    \label{subfig:ssim-score-f}
  \end{minipage}
  
  \caption{This figure illustrates PatchSVD algorithm performance on the Kodak dataset with the patch size of 16 using different score functions, namely, standard deviation (std), mean, and maximum (max). To compare, SSIM, PSNR, and MSE metrics are used. It is demonstrated that standard deviation has a slightly better performance compared to others.}
  \label{fig:score_f_comparison}
\end{figure*}



\section{\uppercase{Results and Discussion}}

\subsection{PatchSVD Performance Based on Patch Size}
\label{sec:various-p-sizes}
Figure \ref{fig:various-patch-sizes} depicts the impact of patch size on the performance of the PatchSVD algorithm across the Kodak and CLIC datasets, as measured by SSIM, PSNR, and MSE metrics. The findings indicate that opting for excessively large patch sizes is not advisable, and the effectiveness of compression may be compromised with patch sizes that are too small, contingent on the characteristics of the dataset. Note that the algorithm falls back to SVD for patch sizes that are too large which is why the plots overlap for some compression ratios.

\subsection{Image Compression Performance Comparison}
To better demonstrate the performance of PatchSVD, we compared the performance of PatchSVD with a fixed patch size against JPEG and SVD on CLIC and Kodak datasets. According to the experiment results in Section \ref{sec:various-p-sizes}, patch sizes 10 and 16 were selected for CLIC and Kodak, respectively. Figure \ref{fig:performance-kodak} illustrates the performance comparison with respect to SSIM, PSNR, and MSE. PatchSVD outperforms SVD in all three metrics on both datasets, although JPEG still performs better than PatchSVD. This is expected because neither of the three metrics are context-aware. Nevertheless, PatchSVD may still be preferred over JPEG in some use cases as explained in Section \ref{sec:artifacts}.

\subsection{Compression Artifacts}
\label{sec:artifacts}
While the compression artifacts of PatchSVD and SVD are usually in the form of colored pixels (sometimes called "stuck pixels", see SVD output in Figure \ref{subfig:artifact-b}), for JPEG, these artifacts take the forms of a general loss of sharpness and visible halos around the edges in the image. In higher compression ratios, the edges of blocks that PatchSVD and JPEG use become visible, too. However, for use cases where sharpness should be maintained locally, PatchSVD is preferable. For example, in Figure \ref{fig:artifacts}, for both samples, the text written on the boat is more legible when the image is compressed with PatchSVD.

\subsection{Choice of Score Function}
We compared various score functions, including taking the maximum value, averaging, and calculating the standard deviation of the pixel values present in the input patch, as shown in Figure \ref{fig:score_f_comparison}. The performance of all the score functions is almost similar, but standard deviation yields better results in terms of SSIM, PSNR, and MSE. The intuition behind this is that standard deviation introduces sensitivity to deviations from the mean which is usually where more complex patterns are present.



\section{\uppercase{Conclusion}}
In this work, we introduced PatchSVD as a non-uniform image compression algorithm based on SVD. Through experiments, we demonstrated that the patch size in the PatchSVD algorithm affects the compression performance. Also, we compared the performance of PatchSVD with JPEG and SVD with respect to SSIM, PSNR, and MSE. We compared the compression artifacts that each algorithm introduced to images and illustrated examples of the cases where PatchSVD was preferable over JPEG and SVD because it produced less destructive artifacts in regions that contained information that would have been lost if we applied standard SVD-based image compression. 
Studying the impact of PatchSVD as an image compression algorithm on the downstream tasks is an interesting future work. Moreover, applying PatchSVD to medical images is a prospective extension because, in medical images, higher resolution is required in the pixels containing diagnostic information compared to the rest of the image and non-uniform local compression could be beneficial. Expert knowledge can lead us to more customized score functions which makes this application even more interesting.

\bibliographystyle{apalike}
{\small
\bibliography{main}}

\end{document}